\documentclass{article}

\usepackage{microtype}
\usepackage{graphicx}
\usepackage{subfigure}
\usepackage{booktabs} 

\usepackage{hyperref}

\usepackage[accepted]{icml2024}

\usepackage{amsmath}
\usepackage{amssymb}
\usepackage{mathtools}
\usepackage{amsthm}

\usepackage[capitalize,noabbrev]{cleveref}

\theoremstyle{plain}
\newtheorem{theorem}{Theorem}[section]

\newtheorem{lemma}[theorem]{Lemma}
\newtheorem{corollary}[theorem]{Corollary}
\theoremstyle{definition}

\theoremstyle{remark}
\newtheorem{remark}[theorem]{Remark}

\usepackage[textsize=tiny]{todonotes}

\usepackage{amsmath,amsfonts,bm}

\def\eqref#1{equation~\ref{#1}}

\def\1{\bm{1}}

\DeclareMathAlphabet{\mathsfit}{\encodingdefault}{\sfdefault}{m}{sl}
\SetMathAlphabet{\mathsfit}{bold}{\encodingdefault}{\sfdefault}{bx}{n}

\DeclareMathOperator*{\E}{\mathbb{E}}

\newcommand{\R}{\mathbb{R}}

\newcommand{\softmax}{\mathrm{softmax}}

\usepackage{amsmath, amssymb, amsthm}
\usepackage{graphicx}
\usepackage{bbm}

\usepackage{hyperref}

\usepackage{pgf, tikz}
\usetikzlibrary{arrows, automata, positioning}

\usepackage{accents}

\usepackage{natbib}
\icmltitlerunning{Q-Probe: A Lightweight Approach to Reward Maximization for Language Models}

\begin{document}

\twocolumn[
\icmltitle{Q-Probe: A Lightweight Approach to\\ Reward Maximization for Language Models}



\icmlsetsymbol{equal}{*}

\begin{icmlauthorlist}
\icmlauthor{Kenneth Li}{harvard,kempner}
\icmlauthor{Samy Jelassi}{cmsa}
\icmlauthor{Hugh Zhang}{harvard,kempner}
\icmlauthor{Sham Kakade}{harvard,kempner}
\icmlauthor{Martin Wattenberg}{harvard}
\icmlauthor{David Brandfonbrener}{kempner}
\end{icmlauthorlist}

\icmlaffiliation{harvard}{John A. Paulson School Of Engineering And Applied Sciences, Harvard University}
\icmlaffiliation{cmsa}{Center of Mathematical Sciences and Applications, Harvard University}
\icmlaffiliation{kempner}{Kempner Institute for the Study of Natural and Artificial Intelligence, Harvard University}

\icmlcorrespondingauthor{Kenneth Li}{ke\_li@g.harvard.edu}
\icmlcorrespondingauthor{David Brandfonbrener}{david\_brandfonbrener@g.harvard.edu}

\icmlkeywords{Machine Learning, ICML}

\vskip 0.3in
]



\printAffiliationsAndNotice{}  

\begin{abstract}
We present an approach called Q-probing to adapt a pre-trained language model to maximize a task-specific reward function. At a high level, Q-probing sits between heavier approaches such as finetuning and lighter approaches such as few shot prompting, but can also be combined with either. The idea is to learn a simple linear function on a model's embedding space that can be used to reweight candidate completions. We theoretically show that this sampling procedure is equivalent to a KL-constrained maximization of the Q-probe as the number of samples increases. To train the Q-probes we consider either reward modeling or a class of novel direct policy learning objectives based on importance-weighted policy gradients. With this technique, we see gains in domains with ground-truth rewards (code generation) as well as implicit rewards defined by preference data, even outperforming finetuning in data-limited regimes. Moreover, a Q-probe can be trained on top of an API since it only assumes access to sampling and embeddings. Code: \url{https://github.com/likenneth/q_probe}.
\end{abstract}

\section{Introduction}
Pre-training on diverse data endows large language models (LLMs) with strong generic language capabilities. However, goal-directed downstream tasks like coding, mathematical reasoning, and dialogue systems require adapting the LLM to the task at hand. 
Since the goals in these tasks can be framed as rewards, this adaptation can take the form of reward maximization.

One approach to do this is finetuning, where the weights of the model are adjusted to improve rewards. Exemplary techniques include reinforcement learning from human feedback (RLHF, ~\citealp{ouyang2022training,rafailov2023direct}) and supervised finetuning on successful examples~\cite{singh2023beyond,dong2023raft,yuan2023scaling}. 


On the other hand, there is evidence that the capabilities required for these downstream tasks have already been learned during pre-training, and the task of adaptation is merely to extract them from the wide spectrum of pre-trained capabilities.
For example, ~\citet{zaken2021bitfit} propose that extremely parameter-efficient finetuning is evidence that the finetuning process is mostly about ``exposing knowledge induced by language-modeling training'', while  \citet{saunders2022self} find that pre-trained language models are usually better at discriminating than generating answers.

Motivated by this line of thought, we present a lightweight approach to reward maximization.
For each downstream task, we keep the whole pre-trained model frozen and only train a small probe which is the same dimension as the residual stream \cite{alain2016understanding}.
We call our method Q-probe as it ``probes'' the expected utility of a completion (action) given a certain prompt (state).

To leverage the Q-probe at inference to generate samples, we perform a sort of rejection sampling.
Specifically, we first draw $k$ sampled completions from the LLM given the input prompt and also store the embedding of each prompt-completion pair.
The Q-probe then predicts a value for each embedding, which determines the logits for the $ k$-way softmax distribution that we use to sample the chosen completion.
In theory, we show that this procedure maximizes the KL-constrained value of the probe as $ k$ tends to infinity.


\begin{figure*}[t]
    \centering
    \includegraphics[width=\linewidth]{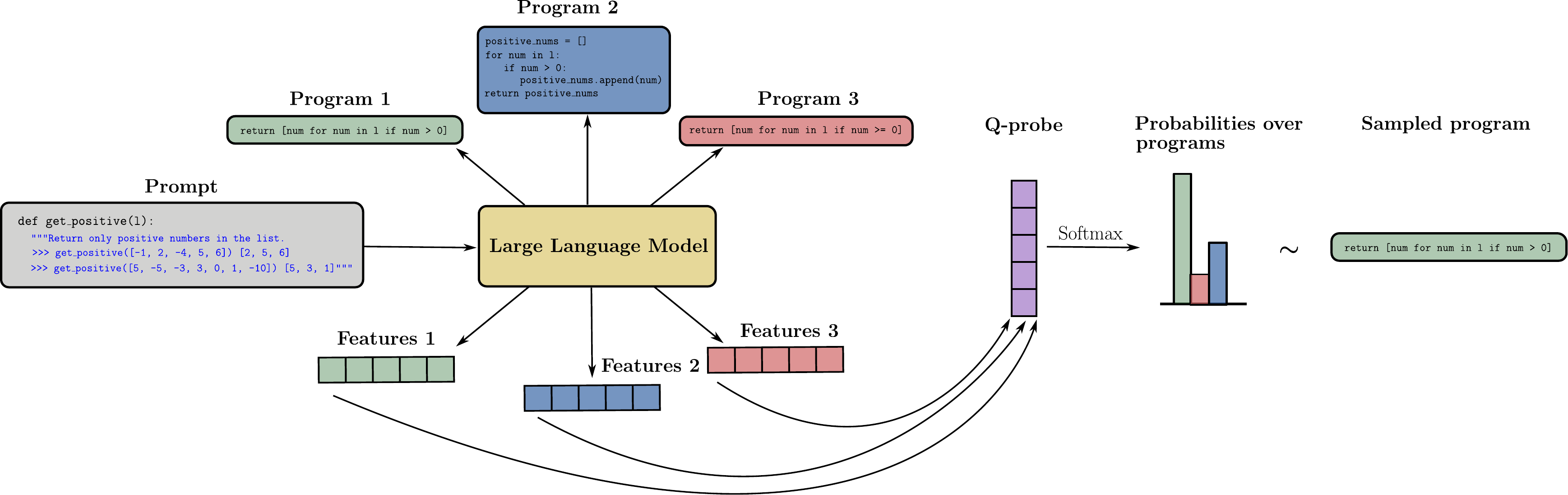}
    \vspace{-7mm}
    \caption{An illustration of the Q-probe inference procedure. Given a prompt, we use the language model to generate $ k=3$ completions (in this case, programs) and the respective embeddings of the $ k$ prompt-completion pairs. Then the linear Q-probe maps the features into the logits of a softmax distribution. We obtain our final sample from the Q-probe by sampling from this distribution.}
    \label{fig:one}
\end{figure*}

First, we evaluate Q-probes with access to ground truth rewards on coding benchmarks our best Q-probe achieves $17\%$ higher accuracy on MBPP~\citep{austin2021program} compared to the base Code-LLaMA-7B \citep{roziere2023code} and outperforms finetuning on successes with LORA \citep{hu2021lora} and few shot prompting. 
Although again, we emphasize that Q-probes are not mutually exclusive with these other techniques and can be combined for even better results.
One key component of the results is a novel objective for training the Q-probes via direct policy learning. 
We find that rather than training the Q-probe to model the rewards, it is more effective to use an importance weighted policy gradient objective.
Since we only need access to samples and embeddings we can train Q-probes on API-based models where gains are more modest ($3\%$ improvement over base model) due to a stronger base model and lack of access to internal model embeddings.

Next, we evaluate Q-probes on learning from human preferences.
We conduct a standardized comparison~\citep{ethayarajh2022understanding} and find that Q-probe outperforms offline PPO and DPO by $6\%$ in terms of win rate as judged by GPT4. 
Moreover, we show that a Q-probe can be trained on top of a KTO finetuned model and outperforms either method individually by an additional $4\%$.
This demonstrates how Q-probes can be combined effectively with other adaptation strategies.


Finally, in terms of computational cost, we should note that using a Q-probe requires substantially less training compute, but more inference-time compute when compared to finetuning.
In our experiments, we can train a Q-probe in a few seconds (since it is just a 4096-dimensional linear model) whereas even parameter efficient finetuning \citep{hu2021lora} takes several hours.
But, at inference, we draw $ k $ samples from the base model rather than 1 from a finetuned model, although improvements in parallel and speculative decoding are making batched decoding easier~\citep{fang2021turbotransformers,yu2022orca,shen2024superposed}.


\section{Related work}

\paragraph*{Probing.} Q-probes leverage the idea of probing to solve reward maximization problems. This idea builds on prior work that uses probes for understanding the internals of neural networks~\citep{alain2016understanding,belinkov2016probing,li2022emergent}. A probe is a classifier or regressor that takes internal activations of a network as its input and is trained to predict a feature of interest, e.g., part of speech, parse tree depth, or the expected reward in our case.

\paragraph*{Rejection sampling.} Rejection sampling for reward maximization is not a new idea. In fact, \citet{gao2023scaling,ganguli2022red,rafailov2023direct} also evaluate rejection sampling as one of their baselines. However, their selector model is instantiated by the preference language model trained in a similar way to the first stage of RLHF by~\citet{ouyang2022training}. This version of rejection sampling not only involves higher training cost but is also double the inference cost to run the reward model while evaluating the Q-probe is essentially free in comparison to the base model.

\paragraph*{Rejection sampling + finetuning.}
Another line of work finetunes or distills models on top of data that is acquired by rejection sampling \citep{singh2023beyond, dong2023raft, yuan2023scaling, rafailov2023direct}. In this work, we just focus on a lightweight way to do the rejection sampling, but adding some sort of distillation step on top to reduce inference cost could be an interesting future direction.

\paragraph*{Iterative finetuning.}
While we focus our experiments primarily on offline settings for simplicity, there is also an interesting direction 
Iterative finetuning for reward maximization. \citep{anthony2017thinking, gulcehre2023reinforced, singh2023beyond, zelikman2022star, dong2023raft}.
The Q-probe idea could be applied inside of iterative algorithms like these and that is an interesting direction for future work.

\paragraph*{Prompting.} An important line of training-free adaptation methods centers around prompting~\citep{salewski2023context} which includes in-context learning (ICL, ~\citealp{min2022rethinking}) and Chain-of-thoughts (CoT, ~\citealp{wei2022chain}). Though it enjoys great flexibility,~\citet{mosbach2023few} reveal by a closer examination that finetuning still outperforms prompting methods. Prompting could also be sensitive to prompt engineering~\citep{lu2021fantastically} and takes up a valuable context window, limiting the amount of data we can feed into it for a fair comparison with Q-probe and finetuning.

\paragraph*{Prompting with reward access.} There are also a host of other inference-time techniques designed for coding and reasoning settings~\citep{zhou2023language,shinn2023reflexion,yao2023tree}. However, they require access to the feedback from the environment at test time which is different from the one-pass setting considered by us.



\section{Setting}

We consider a generic framing that examines downstream language tasks as reward maximization problem. 
In this setting, prompt strings $ x $ are sampled i.i.d. from some distribution $ \mathcal{P}_{prompt}$. 
Then, our model generates completion strings which we will denote by $ a$ (``actions'' in the reinforcement learning lingo).
The goal is to generate completions to maximize some reward function $ r(x,a)$.

Within this setting, we will consider a variety of feedback types (oracle rewards or preferences) as well as interaction levels (offline data or online reward access) that Q-probe can tackle. 
We also only need limited black-box access to the base model.
This section formalizes all of these assumptions about the setting.

\subsection{Feedback: oracle rewards and preferences}

\paragraph{Oracle reward function feedback.} In this setting, we assume access to a train set of prompts $ x \in D_{train}$ and access to the ground-truth or ``oracle'' reward function on the train prompts $ r(x,a)$ for $ x \in D_{train}$ and any $ a$. For example, in coding problems this is assuming that we have test cases for the train prompts.   
For evaluation, we assume access to a test set of prompts $ x \in D_{test}$ and also the reward function on the the test prompts.

The goal when given oracle reward feedback is to learn a policy $ \pi$ to maximize expected return:
\begin{align}
    J(\pi) = \E_{x} \E_{a\sim \pi|x} [r(x,a)]
\end{align}

Note, there is a large literature of prior work on using reinforcement learning directly to finetune language models when given access to oracle reward functions, e.g., for single turn language tasks \citep{schulman2017proximal, snell2022offline, ramamurthy2022reinforcement, chang2023learning} or in multiturn settings \citep{zhou2023sotopia,abdulhai2023lmrl}. In contrast, we focus on a lighter weight approach that only requires training probes, but shows how probe training can approximate traditional RL objectives.

\paragraph{Preference feedback.} This is the same as above, except that we have access to pairwise comparisons. For an $ x \in D_{train}$ for any pair of actions $ (a_0, a_1)$ we can get a label $ l \in \{0,1\}$ indicating which action is preferred \citep{christiano2017deep, ouyang2022training, rafailov2023direct} .

The goal when given preference feedback is to learn a policy $ \pi$ that generates actions to maximize the hidden reward function that induces the preferences (if we assume e.g. a Bradley-Terry model of preferences \citep{Bradley1952RankAO}).

\subsection{Online vs. offline access to feedback}

We always assume a fixed dataset of contexts (i.e. prompts) $ x_i \in D_{train}$. For example, these could be programming problems, math questions, or user queries. From these prompts, we consider two possible levels of access to the feedback source:
\begin{enumerate}
    \item Online. With online access, we can query the reward or preference of \emph{any} action $ a$ or action pair $ (a_0, a_1)$ from any context $ x_i$ in the training set to get $ r(x_i, a)$. This setting is reasonable if we have unit tests for programming or a human in the loop for preference learning. 
    \item Offline. In the offline setting, we assume that the dataset also contains actions or action tuples and reward or preference labels. So the data has tuples of $ (x_i, a_i, r_i) $ or $ (x_i, (a_0)_i, (a_1)_i, y_i)$. We can only access the rewards or preferences through these labels and cannot make arbitrary queries.
\end{enumerate}

Our method can function in the offline setting where the dataset is sampled from the base model or in the online setting when only given sampling access to the base model. Throughout the paper we will default to the \emph{offline} setting so that we can learn from fixed datasets.

Our setting is different than other online settings in which the learner can query $ r(x,a)$ at any $ x$ as well as any $a$. For our purposes, we assume that this level of access is too strong since it allows for searching against the reward function on the test set. Examples of methods in this setting are Reflexion \citep{shinn2023reflexion} or LATS \citep{zhou2023language}.
This is an interesting setting, but beyond the scope of this paper and not directly comparable to our results.

\subsection{Access to the LLM}
We assume access to a pretrained language model that gives us two things:
\begin{enumerate}
    \item Sampling from the LM distribution $ p_0$. Given a context $ x $ we can sample a completion $ a$ from $ p_0(\cdot|x)$. 
    \item Access to embeddings. We can extract an embedding $ \phi(x,a) $ of the joint prompt-completion string.
\end{enumerate}

We do not in general assume access to the underlying model to allow for finetuning, and our method will not require such access, but we consider such methods for comparison.
We also do not assume access to densities or logits from the underlying model.
With these assumptions, our method is applicable on top of API-based models.

We are not aware of prior work on learning algorithms that use this access model. So, we will compare to a few baselines that either get more access (full finetuning of open source models) or less access (just sampling with different prompts).

\section{Inference using Q-probes}



\subsection{Defining the Q-probe policy}
To define the Q-probe policy we reweight samples from the base model using a value function. Let $ Q_\theta: \mathcal{X} \times \mathcal{A} \to \R$, then our policy $ \pi_{\theta,k}$ is defined by the following procedure:
\begin{enumerate}
    \item Sample $ a_i \sim p_0|x, \quad 1 \leq i \leq k$.
    \item Sample $a \sim \softmax \left(\left\langle \frac{Q_\theta(x, a_1)}{\beta}, \dots, \frac{Q_\theta(x, a_k)}{\beta} \right\rangle\right)$.
\end{enumerate}
Note that $ Q_\theta$ does not have to represent a Q function in the lingo of RL, and can be any real valued function, this is just a way to define a policy.

\subsection{Theoretical motivation for the Q-probe policy}

To motivate the Q-probe policy, it is instructive to consider the limit as we take $ k \to \infty$. In particular, we will show that in this limit, the policy converges to the optimal KL-constrained policy that maximizes the expected value of the probe $ Q_\theta$. 

\begin{theorem}\label{lem:limit}
    Our policy approaches the following limit
    \begin{align}
        \lim_{k \to \infty} \pi_{\theta,k} (a|x)  = p_0(a|x) \frac{ \exp(Q_\theta(x,a) / \beta)}{\E_{b \sim p_0|x}[\exp(Q_\theta(x,b)/\beta )]} .\nonumber
    \end{align}
\end{theorem}

\begin{corollary}\label{lem:optimal}
    The limiting policy is the KL regularized policy that optimizes the Q-values:
    \begin{align}
        \lim_{k \to \infty} \pi_{\theta,k}  = \arg\max_\pi \E_{a\sim \pi|x} [Q_\theta(x,a)] - \beta \text{KL}(\pi \| p_0)\nonumber
    \end{align}
\end{corollary}

See proofs in Appendix \ref{app:proofs}.

\paragraph{Connection to rejection sampling.} Our softmax sampling algorithm has a clear analogy to more standard rejection sampling. To define the rejection sampling analog, assume we know a value $ M$ such that $ M \geq \exp(Q_\theta(x,a)/\beta)$ for all $a$. Now the algorithm is:
\begin{enumerate}
    \item Sample $ a $ from $ p_0(\cdot|x)$
    \item Accept $ a $ with probability $ \frac{\exp(Q_\theta(x,a)/\beta)}{M}$, otherwise return to step 1.
\end{enumerate}
The runtime to get a sample accepted is $ M $ iterations in expectation. We can view the softmax version as an approximation of rejection sampling with $ k$ in place of $ M$. This gives us consistent runtime and parallelization, but does mean that for finite $ k $, we are only approximately sampling from the target distribution.

This also makes it clear that to send $ \beta \to 0$ we need to send $ M \to \infty$  (and implicitly $ k \to \infty $).

\section{Training algorithms for Q-probes}

So far we have defined the procedure for sampling from a Q-probe policy and shown that this is a reasonable policy definition. Now we move on to demonstrating the variety of learning algorithms that can be used to train the Q-probes. Essentially, we can either attempt to learn reward/value functions or to learn policies directly. Moreover, we can apply this idea to either reward feedback or preference feedback.

\subsection{Learning from oracle reward feedback}

\paragraph{Reward learning.}
The simplest approach is to simply use mean squared error to learn a $ Q $ probe to approximate the oracle reward function directly.
\begin{align}
    L_Q(\theta) = \E_x \E_{a\sim p_0|x} [(Q_\theta(x,a) - r(x,a))^2]
\end{align}
This learned $ Q_\theta$ then induces a policy $ \pi_{\theta, k}$. Note that in the problems we consider, there is only one step of interaction with the environment so the reward function is equal to the $ Q $ function in the RL sense, this is why we call it a Q-probe.

In many of the problems we consider, the rewards are either 0 or 1. In this case we can also estimate the reward with a classification loss like cross entropy (CE). Then the loss is:
\begin{align}
    L_{CE}(\theta) = \E_x \E_{a\sim p_0|x}[&r(x,a) \log \sigma(Q_\theta(x,a)) + \\ &(1-r(x,a)) \log (1 - \sigma(Q_\theta(x,a)))] \nonumber
\end{align}
This learned $ Q_\theta$ also induces a policy $ \pi_{\theta, k}$ in the same way.

\paragraph{Direct policy learning.}
One benefit of Q-probes is that we can derive a loss that more directly tries to optimize the expected return of the policy. For notational convenience, define $ f(a) = \exp( Q_\theta(x,a)/\beta)$. Then we can define the softmax probability as:
\begin{align}
    \rho_\theta(a, \{a_i\}_{i=2}^k) &= \frac{f(a)}{f(a) + \sum_{i=2}^{k} f(a_i)}.
\end{align}
This $ \rho_\theta$ is the probability of sampling $ a $ conditioned on the $ k $ samples from step 1 of the sampling procedure being $ a, a_2, \dots, a_k$. The nice thing about $ \rho_\theta$ is that it approximated the ratio of densities between $ \pi_{\theta, k}$ and $ p_0$. This allows us to define the following importance weighted policy gradient loss:
\begin{align}
    L_{PG}(\theta) &= \E_{x}\E_{a\sim p_0|x} \left[ - r(x,a) \frac{\pi_\theta^k(a|x)}{p_0(a|x)}\right] \\
    &\approx \E_x \E_{ \substack{a \sim p_0|x\\ a_2, \dots, a_k \sim p_0|x} } \left[ - r(x,a) \rho_\theta(a, \{a_i\}_{i=1}^k)\right]\nonumber
\end{align}
Where by \cref{lem:limit} we have that this approximation is exact as $ k \to \infty$.

As is standard in the policy gradient literature, we can also introduce a baseline $ b(x)$ and replace $ -r(x,a)$ in the loss by $ -(r(x,a) - b(x))$ \citep{greensmith2004variance, schulman2015trust}. In practice, we use the context-independent mean reward in the dataset as our baseline.

\begin{remark}
    This PG loss ends up looking much like a contrastive loss, which has traditionally been used for representation learning \citep{wu2018unsupervised,oord2018representation}. Here, the contrastive loss arises naturally since the inference-time procedure of selecting one sample from many requires us to compare and contrast a set of samples. By directly tying the loss to the inference procedure we can force the model to allocate its errors in such a way that performs better when selecting a sample by softmax.
\end{remark}

\subsection{Learning from preference feedback}

\paragraph{Reward learning.}
The simplest approach to use Q-probes to learn from preferences is to use the probe to learn a reward model using a Bradley-Terry model. The per sample loss is:
\begin{align}
    \ell(x, a_w, a_l, \theta) = \sigma (Q_\theta(x, a_w) - Q_\theta(x, a_l))
\end{align}
And the full $ Q$-preference loss function becomes:
\begin{align}
    L_{QP}(\theta) = \E_{\substack{x \\ a_w, a_l \sim p_0}} [- \log \ell(x, a_w, a_l, \theta)]
\end{align}
This learned $ Q_\theta$ then induces a policy $ \pi_{\theta, k}$.

\begin{remark}
    The preference learning reward objective has a sort of contrastive flavor as well. Since we pair positive and negative samples and incentivize giving them different values, this loss matches better with the downstream inference procedure of sampling many completions and choosing one. 
\end{remark}

Finally, while we did not find it to be useful in practice, it is also possible to parameterize direct policy learning objectives from preference feedback with Q-probes as in DPO \citep{rafailov2023direct}. A full derivation can be found in \cref{app:preference}.

\section{Oracle reward experiments}

For our first experiment, we evaluate the ability of Q-probes to maximize ground-truth oracle rewards. 
Specifically, we focus on a program synthesis as a task with oracle rewards given by evaluating test cases. 
We train probes using the training set from MBPP~\citep{austin2021program} and test on the MBPP test set as well as evaluating generalization to HumanEval~\citep{chen2021evaluating}.
To see if the method generalizes to mathematical capabilities, we carry out experiments on GSM-8K in~\autoref{app:gsm}.

Rather than using a raw LLM as the base model, we start from a model that has already been finetuned on coding data \citep{chen2021evaluating, roziere2023code, li2023starcoder, azerbayev2023llemma}. This supervised finetuning facilitates more effective Q-probing for task-specific rewards.
Specifically, we present two sets of results, first building on top of Code-LLaMA-7B \citep{roziere2023code} and second building on top of the OpenAI API to demonstrate how Q-probes can be applied to API models.

\subsection{Setup}

We train models on the MBPP train set which consists of 464 programming prompts with test cases. We consider the reward to be 1 if all tests are passed and 0 otherwise. For each training prompt, we can generate as many completions as we want from the base model to automatically label with these rewards.
We sample from the base model with temperature 0.8 and top-p 0.95, following \citep{roziere2023code}, unless otherwise noted. For experiments on Code-LLaMA-7B, we take the $26$th hidden layer of the same model for embeddings\footnote{Probe performance usually peaks at an intermediate layer~\citep{hewitt2019structural}.}. For OpenAI API experiments, we experiment with both embedding API calls as well as Code-LLaMA-70B. Unless otherwise stated, the Q-probe is a 1-layer (linear) probe, the optimizer is Adam \citep{kingma2014adam}, the learning rate is $ 5e-5 $, the batch size is 1000, and we train for 150 epochs.
For the PG loss, we need multiple samples from one prompt to compute the loss. To do this, we group samples by prompt and reshape the batch so it contains 100 problems with 10 samples from each problem. 

We evaluate the models on the MBPP test set of 500 programming prompts with test cases and also test generation to HumanEval dataset which has 164 prompts with test cases. The HumanEval dataset has a slightly different format, but contains problems of a similar level of difficulty to test the generalization abilities of the probes. 

\begin{table}[t]
\caption{Expected return for Q-probes on top of Code-LLaMA-7B, trained on 464 problems from MBPP-train. For Q-probe inference we use $ k = 48$ and $ \beta=0.1$. Q-probe results are the mean over 10 training runs.}
\label{tab:mbpp-llama}
\vskip 0.15in
\begin{center}
\begin{small}
\begin{sc}
\begin{tabular}{lcccr}
\toprule
Method & MBPP-Test & HumanEval \\
\midrule
Baseline (Pass@1) & 0.29 & 0.24 \\
Baseline (Greedy) &  0.38 & 0.30 \\
5-shot on successes & 0.42 & 0.33 \\
SFT on successes & 0.42 & 0.32 \\
Prompt RM & 0.31 & 0.25 \\
Finetune RM & 0.34 & 0.26 \\
\midrule
Q-probe $L_Q$ & 0.38 & 0.29\\
Q-probe $L_{CE}$ & 0.40 & 0.32 \\
Q-probe $ L_{PG}$ & 0.46 & 0.34\\
\midrule
5-shot + Q-probe $ L_{PG}$ & \textbf{0.52} & \textbf{0.39}\\
\midrule
(Skyline) Pass@48 & 0.76 & 0.77\\
\bottomrule
\end{tabular}
\end{sc}
\end{small}
\end{center}
\vskip -0.1in
\end{table}

We consider a variety of baselines. First, we report the average success rate of the base model with default temperature sampling (\textsc{Baseline (Pass@1)}). We also report greedy sampling from the base model (\textsc{Base-Greedy}). We include a few-shot baseline where we sample 5 successful completions from the training dataset and put them into context and then sample with temperature 0 (\textsc{5-shot on successes}). 
We also include a skyline of pass@48 which has oracle access to the ground truth rewards.

For the Code-LLaMA model, we have white-box access to the model so we also add baselines that use LORA finetuning \citep{hu2021lora}. We consider supervised finetuning on the successful completions from the training data followed by greedy decoding (\textsc{SFT on successes}) \citep{singh2023beyond, dong2023raft}. We also consider two kinds of rejection sampling alternatives: one using instruction to prompt the model to judge its own generation (\textsc{Prompt RM}) and the other using a LORA finetuned reward model instead of a lightweight probe (\textsc{Finetune RM}). For the latter, we add linear probe to the base policy model at the last residual steam; but different to Q-probe, the whole model is tuned for judging reward with Lora~\citep{hu2021lora}. At inference time, both rejection sampling baselines adopt hardmax over 48 generations.

\subsection{Code-LLaMA results} 

We present results for training Q-probes on top of Code-LLaMA-7B in \cref{tab:mbpp-llama}. The main finding is that Q-probe with the policy gradient loss $ L_{PG}$ is the best model. This confirms the idea that finding a loss that is a more direct proxy for the downstream task leads to better outcomes. 
The policy gradient loss contrasts many samples for the same prompt, which mirrors the inference procedure and leads to better performance at test time.

Also, recall that Q-probe is easily combined with other methods. To illustrate this, we combine few shot prompting with Q-probe. This leads to even better performance, showing how the different inference procedures are actually leading to complementary improvements in performance that neither approach achieves on its own.

At a higher level, it is also important to note the benefits of training such small and lightweight probes. Because the probe is so small, we can extract a useful discriminator from the generative model with only a small amount of training and use this probe to improve performance.

\begin{figure}[t]
    \centering
    \includegraphics[width=\linewidth]{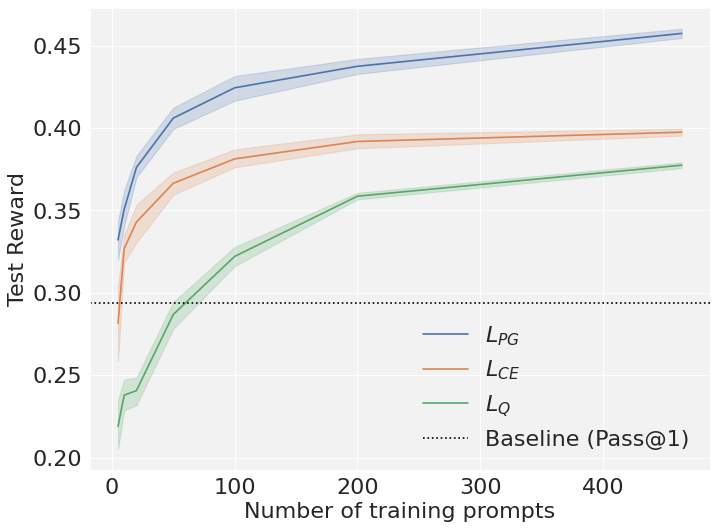}
    \vspace{-0.5cm}
    \caption{How MBPP test reward scales with the size of the training dataset. At inference we fixing $ K = 48$ and $ \beta = 0.1$. Error bars show $95\%$ confidence interval over 10 training runs.}
    \label{fig:mbpp_data}
\end{figure}

\cref{fig:mbpp_data} shows how the Q-probes scale as we vary the number of prompts in the training dataset. In this experiment we take 10 different random samples of $ n $ prompts and train Q-probes on a dataset of completions of these prompts from the base model. We find that The PG loss consistently beats the Q and CE losses and that data efficiency can be quite good, achieving 0.4 test reward from only 50 prompts.

\begin{figure}[h]
    \centering
    \includegraphics[width=\linewidth]{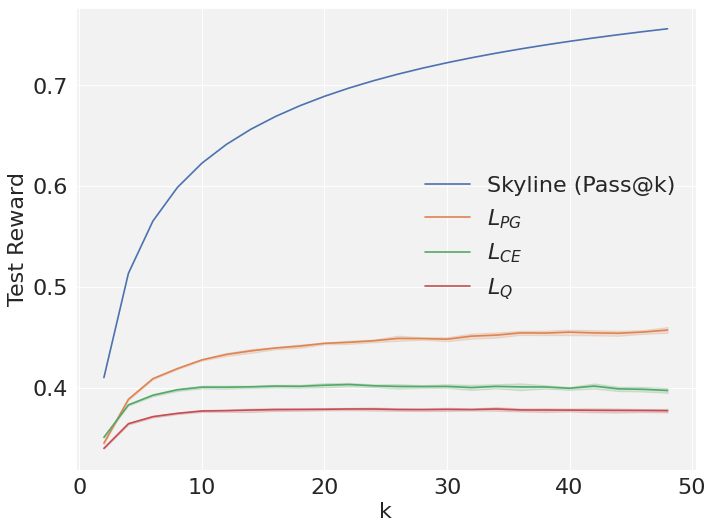}
    \vspace{-0.5cm}
    \caption{How MBPP test reward scales with inference-time compute when sweeping over $K$ with $ \beta =0.1$. Error bars show $95\%$ confidence interval over 10 training runs.}
    \label{fig:mbpp_k}
\end{figure}

\cref{fig:mbpp_k} shows how the Q-probes scale as we vary $ k $, the number of samples drawn at inference time. We see that the model trained with PG loss sees consistent improvement with $ k$, although it is beginning to saturate. In contrast, $ L_{Q}$ and $ L_{CE}$ actually see performance slightly degrading as we increase $ k$. This again affirms how matching the training loss to the inference procedure is beneficial.


\begin{figure}[h]
    \centering
    \includegraphics[width=\linewidth]{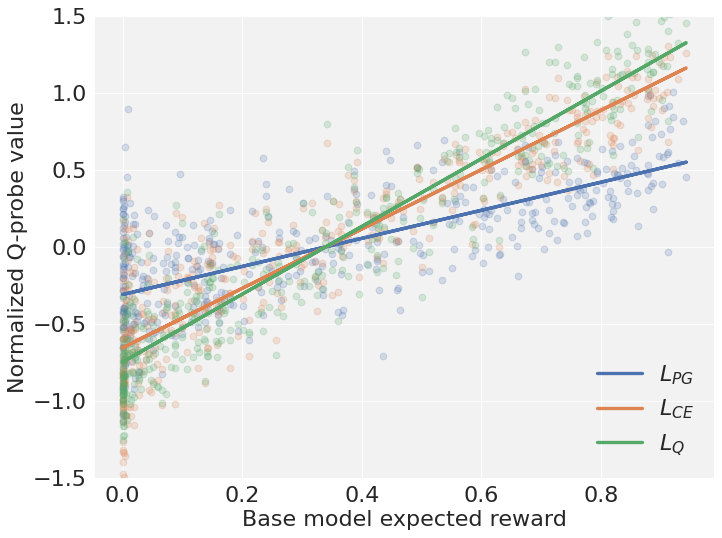}
    \vspace{-0.5cm}
    \caption{Per-problem correlation between base model expected reward and Q-probe value (centered and normalized by standard deviation). Each point corresponds to a prompt in the training set and averages across the 200 sampled completions. $L_{PG}$ learns by contrasting completions to the same prompt, so it learns a probe that is less prompt-dependent.}
    \label{fig:corr}
\end{figure}

Finally, \cref{fig:corr} attempts to provide some intuition about how $ L_{PG}$ differs from $ L_Q$ and $ L_{CE}$ in a way that is beneficial. First and foremost $ L_{PG}$ attempts to optimize a proxy of the test metric, expected reward. This experiment tries to look at a lower level to see how this changes the learned models. 
Intuitively, $ L_Q$ and $L_{CE}$ treat samples $ a $ from the same $ x $ independently (since they just sum over all samples) and end up allocating a good amount of capacity to classifying which \emph{prompts} are hard (causing higher slope in the figure). But the $ L_{PG}$ loss forces the model to learn which completions are good when compared to each other for the \emph{same} prompt. The contrastive nature of this loss helps the model allocate capacity more effectively to the part of the problem that matters: comparing different completions of the same prompt.

\begin{table}[h]
\caption{Expected return for Q-probe models on top of \texttt{gpt-3.5-turbo-1106} and \texttt{CodeLlama-70b-Python} embeddings.
Q-probe inference uses $ k = 48$ and $ \beta = 0.1$. Q-probe results are the mean over 10 training runs.}
\label{tab:mbpp-openai}
\vskip 0.15in
\begin{center}
\begin{small}
\begin{sc}
\begin{tabular}{lcccr}
\toprule
Method & MBPP-Test & HumanEval\\
\midrule
Baseline (Pass@1) & 0.65 & 0.54 \\
Baseline (Greedy) &  0.65 & 0.59 \\
5-shot on successes & 0.66 & 0.61 \\
\midrule    
Q-probe $L_Q$  & 0.68 & 0.57\\
Q-probe $L_{CE}$  & \textbf{0.69} & \textbf{0.64}\\
Q-probe $ L_{PG}$  & \textbf{0.69} & 0.58 \\
\midrule
(Skyline) Pass@48 & 0.80 & 0.81 \\
\bottomrule
\end{tabular}
\end{sc}
\end{small}
\end{center}
\vskip -0.1in
\end{table}

\subsection{OpenAI API results}

Finally, we conduct a similar experiment on top of generations of the OpenAI API. Results are reported in \cref{tab:mbpp-openai}. We use embeddings from \texttt{CodeLlama-70b-Python} since embeddings are not available from the API generative model. 
We find gains over the baselines on both datasets. 

While this is a nice proof of concept that Q-probes can be applied on top of API-based models, the results are not as strong as they were for Code-LLaMA. We hypothesize that this is largely for two reasons: (1) the base model is much stronger on the task and has likely been finetuned to do particularly well at these coding tasks so there is simply less room for reweighting to help, and (2) we do not have access to the embeddings from the model itself and the open source embeddings from Code-LLaMA are likely less performant. 

We also experimented with embeddings from the OpenAI API, and found them to work less well than the Code-LlaMa embeddings. Full results and discussion of these experiments are in~\autoref{app:mbpp-ada}.


\subsection{Additional Experiments on GSM-8K}
\label{app:gsm}

We also conduct experiment on GSM-8K with Code-Llama-7B, $ k = 48$ and $ \beta=0.1$, following the implementation of~\citep{gao2022pal,cobbe2021gsm8k}, using 8-shot evaluation with code adopted from the Code Generation LM Evaluation Harness project~\citep{bigcode-evaluation-harness}. Results in~\autoref{tab:gsm-llama} show a similar trend as the experiments on coding. 

\begin{table}[h]
\caption{Expected return on GSM-8K, trained on 7473 problems from the training set of GSM-8K. Hyperparameters kept the same as~\autoref{tab:mbpp-llama}. Evaluation protocols follow~\cite{gao2022pal}.}
\label{tab:gsm-llama}
\vskip 0.15in
\begin{center}
\begin{small}
\begin{sc}
\begin{tabular}{lccr}
\toprule
Method & GSM-8k \\
\midrule
Baseline (Pass@1) & 0.25\\
Baseline (Greedy) &  0.29 \\
\midrule
Q-probe $L_Q$ & 0.36 \\
Q-probe $L_{CE}$ & 0.43 \\
Q-probe $ L_{PG}$ & \textbf{0.45} \\
\midrule
(Skyline) Pass@48 & 0.80 \\
\bottomrule
\end{tabular}
\end{sc}
\end{small}
\end{center}
\vskip -0.1in
\end{table}

\section{Preference feedback experiments}

We also experiment with Q-probe on learning from human preference data. We follow the set-up and implementation of~\citet{ethayarajh2023halos} strictly unless otherwise specified. We use the combination of three open-source preference datasets---Anthropic Helpfulness and Harmlessness (HH)~\citep{ganguli2022red}, OpenAssistant~\citep{kopf2023openassistant}, and Stanford Human Preferences Dataset (SHP)~\citep{ethayarajh2022understanding}. Experiments are carried out on LLaMA-7B~\citep{touvron2023llama}.

\subsection{Setup}

We first extract features for probe training. Combining the training sets of three datasets together, we obtain a dataset with $200,336$ training pairs, each containing a winning completion and a losing completion. We concatenate the prompt with both completions and run a forward pass of the model to extract embeddings. Note that our Q-probing is applied on the supervised finetuned model, which is also the starting point for the compared methods~\citep{ouyang2022training,rafailov2023direct,ethayarajh2023halos}. Offline PPO, DPO, and KTO use different loss functions to finetune the model weights from this supervised finetuned model.

Upon finishing training, we sample $48$ samples for each prompt in the test set and embed them with the model. The Q-probe then returns the scores for each completion. Here we use $ \beta = 0$ and select the argmax of the scores. During evaluation, the model's completion is compared against the winning completion in the data for that prompt by GPT-4 as the judge to compute the ``win rate''.

\paragraph*{Experiment Details} We implement the Q-probe with a 1-layer probe, trained at a learning rate of $5e-5$ with batch size $1024$ for $150$ epochs using $20\%$ of the whole training set used by other methods, which is $40,067$ pairs of winning and losing generations. All methods use nucleus sampling ($p=0.95$) and temperature $1.0$ at inference~\citep{holtzman2019curious}.

\subsection{Experimental Results}
\begin{figure}
    \centering
    \includegraphics[width=1\linewidth]{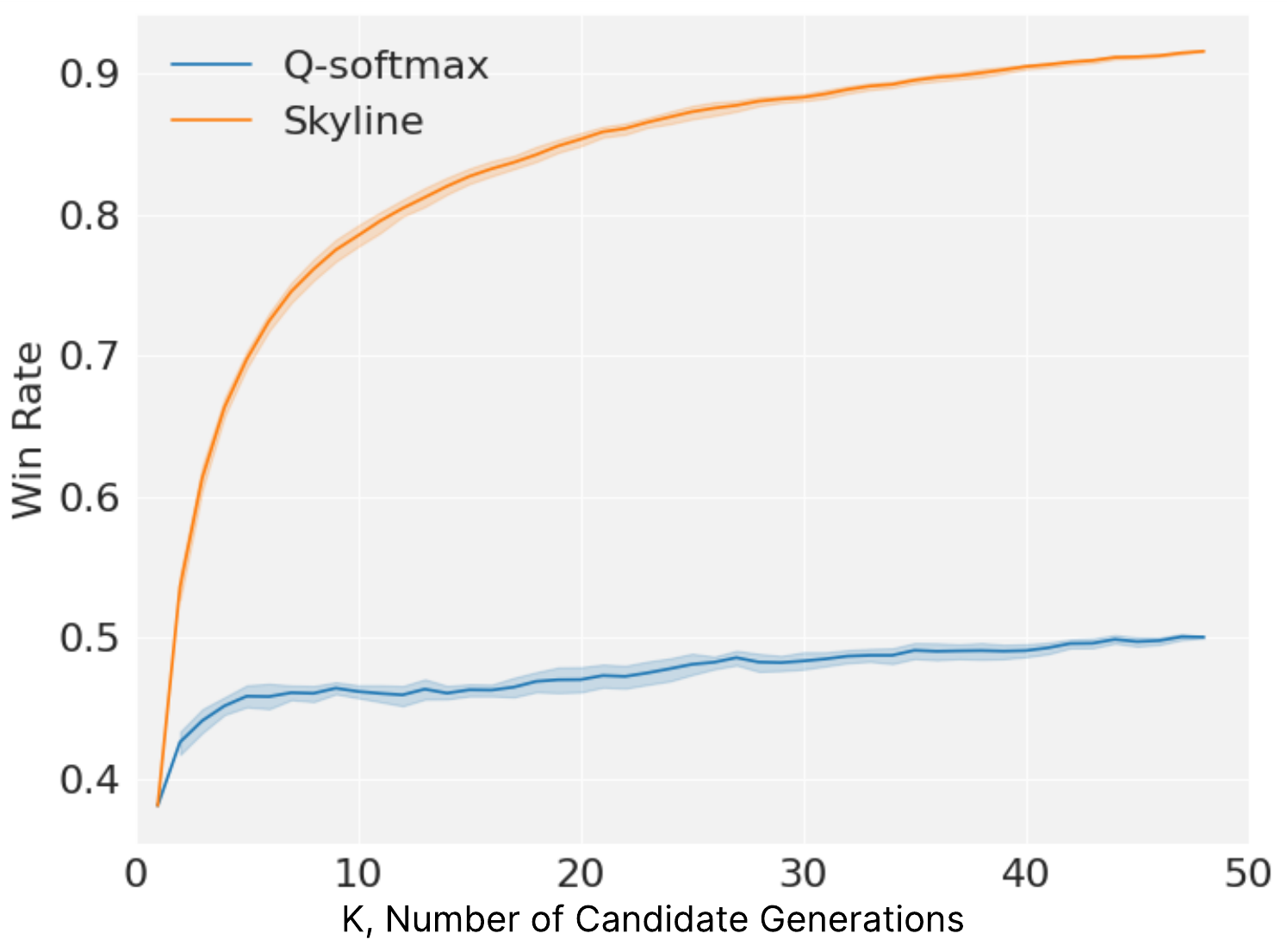}
    \vspace{-8mm}
    \caption{How the win rate of Q-probe scales with inference-time compute on preference learning benchmarks. The skyline shows the performance of a perfect oracle selector. The shaded area represents $95\%$ confidence interval for $10$ runs.}
    \label{fig:rlhf_k}
\end{figure}

\begin{figure}
    \centering
    \includegraphics[width=1\linewidth]{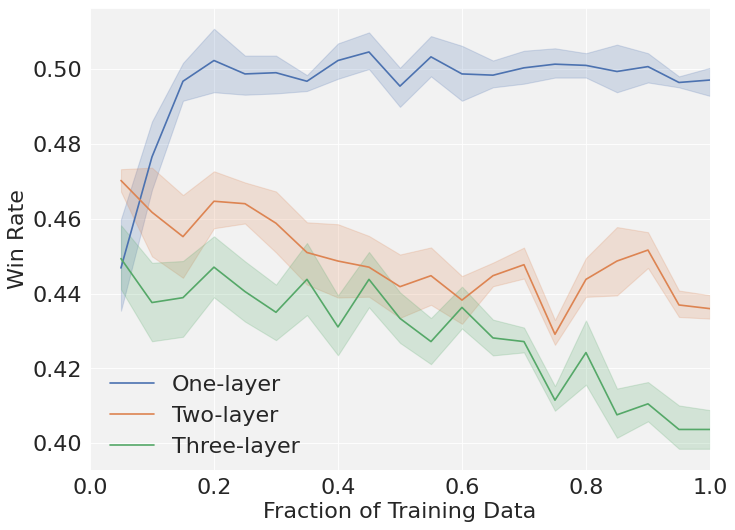}
    \vspace{-8mm}
    \caption{How the win rate on human preference learning benchmarks scales with the percentage of data used for training three different kinds of probes, from $5\%$ to $100\%$ at an interval of $5\%$. There are in total $200,336$ training pairs. }
    \label{fig:layer_scale}
\end{figure}

\begin{table}[h]
\caption{Comparison of different preference learning methods on the combination of three datasets. \cite{ethayarajh2023halos}'s setting is exactly followed; numbers for base models are taken from their paper. The base model is LLaMA-7B after SFT training. }
\label{tab:rlhf}
\vskip 0.15in
\begin{center}
\begin{small}
\begin{sc}
\begin{tabular}{lc}
\toprule
Method & Win Rate ($\%$) \\
\midrule
Baseline & 37.86 \\
PPO (offline) & 44.07 \\
DPO & 44.97 \\
KTO &51.46 \\
\midrule
Q-probe w/ $L_{QP}$  & 50.10 \\
\midrule
KTO + Q-probe w/ $L_{QP}$ &  \textbf{55.01} \\
\midrule
(Skyline) Pass@48 & 91.59 \\
\bottomrule
\end{tabular}
\end{sc}
\end{small}
\end{center}
\vskip -0.1in
\end{table}

\autoref{tab:rlhf} presents our results on human preference data. Starting from the same supervised finetuned model, Q-probe outperforms strong existing methods like PPO (offline) and DPO, while performing on par with KTO. We also experiment with swapping the base model with the KTO-finetuned model, and show that Q-probe on the KTO-finetuned model outperforms either KTO alone or Q-probing on the base model. This shows how our proposed inference-time algorithm is orthogonal to existing finetuning methods and that they can be applied together. 

In~\autoref{fig:rlhf_k}, we vary the amount of inference-time compute by varying the $k$, the number of samples we generate. Improvement begins to plateau around $k=5$ but further scaling continues to slowly increase the win rate. 

In~\autoref{fig:layer_scale}, we examine how much data is required for the Q-probe to work well. For the 1-layer linear probe, thanks to its simplicity, only $20\%$ of the data is required to reach plateaued performance, making Q-probe a worth-considering candidate method when the available preference data is small. We also experiment with more powerful probe architectures, e.g. 2 or 3-layer MLPs, discovering this actually harms performance by overfitting (note that larger datasets also lead to more training since we fix the number of epochs and batch size). In one interpretation, the Q-probe discovers a linear preference direction in the hidden space of the LLM, which could be related to the formation of linear structures in various neural networks~\citep{radford2017learning,voynov2020unsupervised,rogers2021primer}.

\section{Discussion}

We have proposed Q-probe, a lightweight approach to maximize reward on downstream tasks given a pre-trained language model. 
Q-probes can be used effectively as a complement to or replacement for other techniques like finetuning or prompting.
On two settings with access to oracle rewards and human preference pairs respectively, Q-probe outperforms strong baselines. 
For anyone who does not have the resource or access to finetune large language models but wishes to adapt them for their own downstream tasks, Q-probe can serve as a solid replacement, and even given a finetuned model, Q-probe can be added on top to leverage more inference-time compute to squeeze out better performance.

One interesting direction for future work is to study in more depth what sort of probes are learned by Q-probes on different tasks. Are the probes possibly similar across tasks? There could also be interesting connections to ``task vectors''~\citep{ilharco2022editing}.

Finally, Q-probe is inspired by, and corroborates, earlier findings about the generation-discrimination (GD) gap in large language models~\citep{saunders2022self}. This work essentially demonstrates the technical possibility of closing GD gap by rejection sampling---use the stronger discrimination capability to help the weaker generation capability. One interesting direction for future work is to investigate whether fine-tuning with the improved policy could, in turn, enhance the discrimination capability, and if so, how long this self-improving spiral could last.





\subsection*{Impact Statement}

This paper presents work whose goal is to advance the field of Machine Learning. There are many potential societal consequences of our work, none which we feel must be specifically highlighted here. A benefit of the proposed lightweight approach is that it lowers carbon emissions.

\subsection*{Acknowledgments}

Kenneth Li and Hugh Zhang are supported by a fellowship from the Kempner Institute for the Study of Natural and Artificial Intelligence at Harvard University. Kempner Institute computing resources enabled this work. Hugh is additionally supported by a Graduate Research Fellowship from the National Science Foundation. Samy Jelassi acknowledges funding supported by the Center of Mathematical Sciences and Applications. This work has been made possible in part by a gift from the Chan Zuckerberg Initiative Foundation to establish the Kempner Institute for the Study of Natural and Artificial Intelligence. Sham Kakade acknowledges funding from the Office of Naval Research under award N00014-22-1-2377.

\bibliography{references}

\onecolumn
\appendix

\section{Proofs}\label{app:proofs}

\begin{theorem}\label{th:policy_limit}
        Our policy approaches the following limit
    \begin{align}
        \lim_{k \to \infty} \pi_{\theta,k} (a|x)  = p_0(a|x) \frac{ \exp(Q_\theta(x,a) / \beta)}{\E_{b \sim p_0|x}[\exp(Q_\theta(x,b)/\beta )]} 
    \end{align}
\end{theorem}

\begin{proof}
    First note that we can write the density of $ \pi_{\theta, k}$ as follows:
    \begin{align}
        \pi_{\theta, k}(a|x) &= \sum_{\{a_i\}_{i=1}^k \in \mathcal{A}^k} \pi_{\theta, k}(a|x, \{a_i\}_{i=1}^k) p_0(\{a_i\}_{i=1}^k| x)\\
        &= \E_{\{a_i\}_{i=1}^k \sim p_0|x}\left[\pi_{\theta, k}(a|x, \{a_i\}_{i=1}^k) \right]\\
        &= \E_{\{a_i\}_{i=1}^k \sim p_0|x}\left[  \sum_i \mathbb{I} \{a_i=a\} \frac{ \exp( Q_\theta(x,a_i)/\beta)  }{\sum_j\exp(Q_\theta(x,a_j)/\beta)}\right]  \\
        &= \E_{\{a_i\}_{i=1}^k \sim p_0|x} \left[\frac{\sum_i \mathbb{I} \{a_i=a\} }{\sum_j\exp(Q_\theta(x,a_j)/\beta)}\right] \exp( Q_\theta(x,a)/\beta)\\
        &= \E_{\{a_i\}_{i=1}^k \sim p_0|x} \left[\frac{\frac{1}{k} \sum_i \mathbb{I} \{a_i=a\} }{\frac{1}{k}\sum_j\exp(Q_\theta(x,a_j)/\beta)}\right] \exp( Q_\theta(x,a)/\beta)
    \end{align}
    Taking the limit of $ k \to \infty$ and using Law of Large Numbers ($a_i$'s are i.i.d.) 
    \begin{align}
        \lim_{k \to \infty} \pi_{\theta, k}(a|x) &= \lim_{k \to \infty} \E_{\{a_i\}_{i=1}^k \sim p_0|x} \left[\frac{p_0(a|x) }{\E_{b\sim p_0|x}\left[\exp(Q_\theta(x,b)/\beta)\right]}\right] \exp( Q_\theta(x,a)/\beta) \\
        &= p_0(a|x) \frac{\exp( Q_\theta(x,a)/\beta)}{\E_{b\sim p_0|x}\left[\exp(Q_\theta(x,b)/\beta)\right]}
    \end{align}
\end{proof}

\begin{corollary}
    The limiting policy is the optimal KL regularized policy:
    \begin{align}
        \lim_{k \to \infty}\pi_{\theta, k}(a|x) = p_0(a|x) \frac{ \exp(Q_\theta(x,a) / \beta)}{\E_{b \sim p_0|x}[\exp(Q_\theta(x,b)/\beta )]} = \arg\max_\pi \E_{a\sim \pi|x} [Q_\theta(x,a)] - \beta KL(\pi \| p_0)
    \end{align}
\end{corollary}
The proof follows directly from Appendix A.1 in \citet{rafailov2023direct}.

\section{Preference learning objectives}\label{app:preference}

\paragraph{Direct policy learning.} 
Alternatively, we can take inspiration from DPO~\citep{rafailov2023direct} and learn the policy directly. Recall that to define the DPO loss, we consider the per-sample likelihood of an example as:
\begin{align}
    p(x, a_w, a_l, \theta) = \sigma \left(\alpha \frac{\pi_\theta^k(a_w|x)}{p_0(a_w|x)} - \alpha \frac{\pi_\theta^k(a_l|x)}{p_0(a_l|x)} \right)
\end{align}
And then the full DPO loss is:
\begin{align}
    L_{DPO}(\theta) = \E_{\substack{x \\a_w, a_l, a_i \sim p_0}} \left[- \log p(x, a_w, a_l, \theta) \right]
\end{align}
When using Q-probing as the policy, we can use $ \rho_\theta$ to approximate the ratio between $ \pi_{\theta, k}$ and $ p_0$ in the expression for $ p$. To do this, let $ \tilde p(a, a_w, a_l, {a_i}_{i=2}^k, \theta)$ be defined as follows:
\begin{align}
    \sigma \left(\alpha \rho_\theta(a_w, \{a_l, a_i\}_{i=3}^k) - \alpha \rho_\theta(a_l, \{a_w, a_i\}_{i=3}^k) \right)
\end{align}

Then $ L_{DPO}(\theta)$ can be approximated by:
\begin{align}
    \approx \E_{\substack{x \\a_w, a_l, a_i \sim p_0 \\ a_2, \dots, a_k \sim p_0}}\left[- \log \tilde p(a, a_w, a_l, {a_i}_{i=2}^k, \theta) \right]
\end{align}
where again by \cref{lem:limit} this approximation becomes exact as $ k \to \infty$.

We can expand $\rho_\theta$  in the above loss and notice that it becomes:
\begin{align}
    \sigma\left(\alpha \frac{\exp(Q(x, a_w)/\beta) - \exp(Q(x, a_l)/\beta)}{\exp(Q(x, a_w)/\beta) + \exp(Q(x, a_l)/\beta) + \sum_i \exp(Q(x, a_i)/\beta) } \right)
\end{align}
If there are no $ a_i$ we can still implement this with just two samples $ a_w$ and $ a_l$ at which point it begins to look much like the reward modeling loss, but with the softmax incorporated.


\section{Additional OpenAI API Experiments}
\label{app:mbpp-ada}

Here we experiment with embeddings from the OpenAI API. As shown in~\autoref{tab:mbpp-ada}, embeddings from \texttt{text-embedding-3-small} underperforms the Code-LLaMA embeddings and did not yield any performance gains over the baseline models. This is likely because in addition to likely using a smaller, less performant model than gpt-3.5, the API embedding models are likely trained for retrieval applications rather than generation. This difference may harm performance as a Q-probe, but future work is needed to more deeply understand the differences between various embeddings as Q-probes.

\begin{table}[h]
\caption{Expected return for Q-probe models on top of \texttt{gpt-3.5-turbo-1106} and \texttt{text-embedding-3-small}. Q-probe inference uses $ k = 48$ and $ \beta = 0.1$. Q-probe results are the mean over 10 training runs.}
\label{tab:mbpp-ada}
\vskip 0.15in
\begin{center}
\begin{small}
\begin{sc}
\begin{tabular}{lcccr}
\toprule
Method & MBPP-Test & HumanEval\\
\midrule
Baseline (Pass@1) & 0.65 & 0.54 \\
Baseline (Greedy) &  0.65 & 0.59 \\
5-shot on successes & 0.66 & \textbf{0.61} \\
\midrule    
Q-probe $L_Q$  & 0.65 & 0.54\\
Q-probe $L_{CE}$  & 0.65 & 0.54\\
Q-probe $ L_{PG}$  & 0.66 & 0.54 \\
Q-probe $L_Q$ (3 layer) & 0.67 & 0.53 \\
Q-probe $L_{CE}$ (3 layer)  & 0.67 & 0.47\\
Q-probe $ L_{PG}$ (3 layer)  & \textbf{0.68} & 0.51 \\
\midrule
(Skyline) Pass@48 & 0.80 & 0.81 \\
\bottomrule
\end{tabular}
\end{sc}
\end{small}
\end{center}
\vskip -0.1in
\end{table}

\end{document}